\documentclass[a4paper, twoside, 12 pt]{report}

\usepackage[english]{babel}
\usepackage[utf8x]{inputenc}
\usepackage[T1]{fontenc}
\usepackage{setspace}
\usepackage{enumitem}
\usepackage[authoryear]{natbib} 
 
\usepackage{url}

\usepackage{titlesec}
\usepackage{lmodern} 

\titleformat{\chapter}[block]
{\normalfont\huge\bfseries}
{\chaptername\ \thechapter:}
{1em}
{\raggedright}
\titlespacing*{\chapter}{0pt}{20pt}{20pt}

\titleformat{\section}
{\normalfont\Large\bfseries\raggedright}
{\thesection}
{1em}
{}
\titlespacing*{\section}{0pt}{15pt}{10pt}

\titleformat{\subsection}
{\normalfont\large\bfseries\itshape\raggedright}
{\thesubsection}
{1em}
{}
\titlespacing*{\subsection}{0pt}{12pt}{8pt}

\titleformat{\subsubsection}
{\normalfont\normalsize\bfseries\raggedright}
{\thesubsubsection}
{1em}
{}
\titlespacing*{\subsubsection}{0pt}{10pt}{6pt}

\usepackage{tocloft}
\usepackage[a4paper,top=2.5cm,bottom=2.5cm,left=2.5cm,right=2.5cm,marginparwidth=1.75cm]{geometry}
\usepackage{booktabs}
\usepackage{array}
\usepackage{amsfonts}
\usepackage{threeparttable}

\usepackage{pgfgantt}
\usepackage{mathptmx}
\usepackage{xcolor}
\usepackage{amsmath}
\usepackage{graphicx}
\usepackage[colorinlistoftodos]{todonotes}
\usepackage[colorlinks=true, allcolors=blue, pagebackref=true]{hyperref}

\definecolor{imperialblue}{RGB}{173, 216, 230}
\definecolor{impericallightblue}{RGB}{152,189,217}

\title{A Hierarchical Framework for Humanoid\\[0.5 cm] Locomotion with Supernumerary Limbs}
\author{Bowen Zhi}

\begin{document}
\begin{titlepage}

\newcommand{\HRule}{\rule{\linewidth}{0.5mm}} 

\newcommand{\quickwordcount}[1]{%
  \immediate\write18{texcount -1 -sum -merge -q #1.tex output.bbl > #1-words.sum }%
  \input{#1-words.sum}
}


\includegraphics[width=8cm]{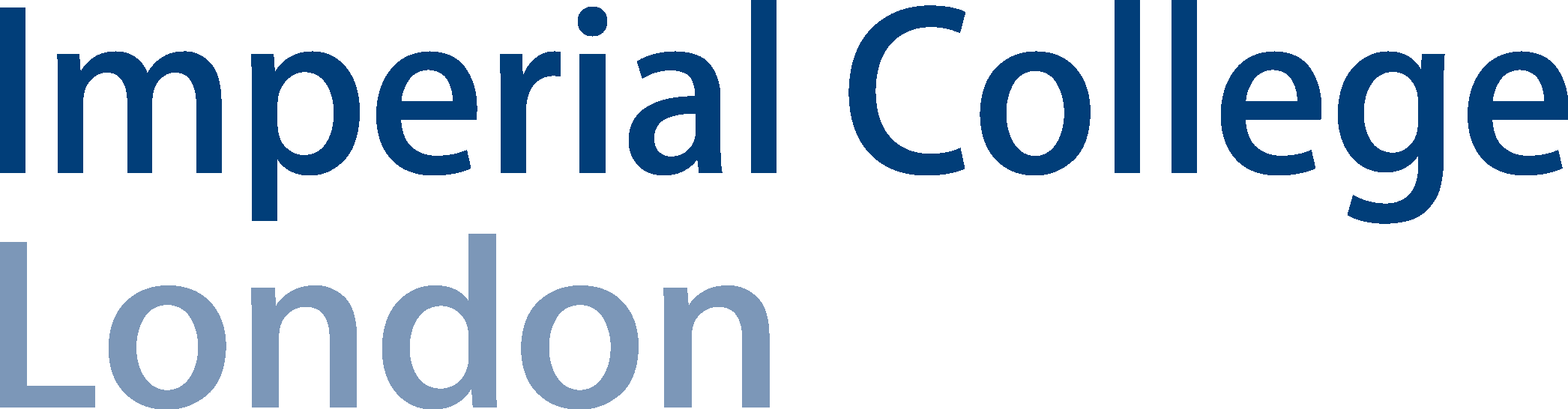}\\[2.2cm] 
 

\center 

\textsc{\LARGE Imperial College London}\\[2.5cm] 

\makeatletter
\HRule \\[0.4cm]
{ \huge \bfseries \@title}\\[0.4cm] 
\HRule \\[3 cm]

\textsc{\large Author: Bowen Zhi}\\[5 cm]

\makeatother


{\large September 10, 2025}\\[2cm] 


\vfill 

\end{titlepage}
\pagenumbering{roman}

\renewenvironment{abstract}{%
    \clearpage
    \vspace*{0.8cm}
    \begin{center}%
        \vspace{10pt}
        \bfseries\Large Abstract
    \end{center}%
    \quotation
}
{%
    \endquotation
}

\begin{abstract}
The integration of Supernumerary Limbs (SLs) on humanoid robots poses a significant stability challenge due to the dynamic perturbations they introduce. This thesis addresses this issue by designing a novel hierarchical control architecture to improve humanoid locomotion stability with SLs. The core of this framework is a decoupled strategy that combines learning-based locomotion with model-based balancing. The low-level component consists of a walking gait for a Unitree H1 humanoid through imitation learning and curriculum learning. The high-level component actively utilizes the SLs for dynamic balancing. The effectiveness of the system is evaluated in a physics-based simulation under three conditions: baseline gait for an unladen humanoid (baseline walking), walking with a static SL payload (static payload), and walking with the active dynamic balancing controller (dynamic balancing). Our evaluation shows that the dynamic balancing controller improves stability. Compared to the static payload condition, the balancing strategy yields a gait pattern closer to the baseline and decreases the Dynamic Time Warping (DTW) distance of the CoM trajectory by 47\%. The balancing controller also improves the re-stabilization within gait cycles and achieves a more coordinated anti-phase pattern of Ground Reaction Forces (GRF). The results demonstrate that a decoupled, hierarchical design can effectively mitigate the internal dynamic disturbances arising from the mass and movement of the SLs, enabling stable locomotion for humanoids equipped with functional limbs. Code and videos are available here: https://github.com/heyzbw/HuSLs.
\end{abstract}

\newpage

\tableofcontents
\clearpage
\pagenumbering{arabic}

\chapter{Introduction}
\label{chap:introduction}

The pursuit of creating versatile humanoid robots capable of operating effectively in human-centric environments represents a grand challenge in robotics. A key determinant of this versatility is the ability to maintain stable bipedal locomotion, a task of significant engineering complexity due to the underactuated and inherently unstable dynamics of legged systems \citep{Collins2005}. This challenge is magnified when humanoid platforms are augmented with additional functional components, such as Supernumerary Robotic Limbs (SLs). SLs promise to dramatically enhance a robot's capabilities, allowing it to perform complex manipulation, carrying, and support tasks that would otherwise be impossible \citep{Parietti2014}. However, the integration of heavy, articulated SLs, such as those capable of carrying payloads up to 30kg, introduces substantial and continuous dynamic perturbations to the main body, severely compromising the stability of the underlying locomotion.

\section{Related Work}
\label{sec:related_work}

The control of bipedal locomotion and the use of auxiliary limbs for stability have been active areas of research, each with a rich history and distinct methodologies.

\subsection{Control of Humanoid Locomotion}
Traditional approaches to humanoid walking have predominantly relied on model-based control, with the Zero-Moment Point (ZMP) criterion being a cornerstone for decades \citep{Kajita2003}. These methods generate dynamically stable trajectories by ensuring the ZMP remains within the support polygon of the feet. While effective in structured environments, ZMP-based controllers often struggle with uneven terrain and in the presence of unforeseen external disturbances, as they rely heavily on precise models and predefined contact schedules.

More recently, Deep Reinforcement Learning (DRL) has emerged as a powerful paradigm for generating sophisticated locomotion gaits without requiring an explicit dynamics model. DRL has proven highly effective for controlling complex legged systems, enabling skills like quadrupedal locomotion over challenging terrain \citep{Lee2020}. A key enabler for this progress has been the development of high-fidelity physics simulators, such as MuJoCo, which provide the vast amounts of data needed for training these policies \citep{Todorov2012}. By leveraging techniques like imitation learning, as demonstrated by the DeepMimic framework, DRL agents can learn complex, physics-based character skills from motion capture data, producing natural and dynamic movements \citep{Peng2018, AlHafez2023}. Algorithms like Proximal Policy Optimization (PPO) have become standard for these tasks due to their stability and data efficiency \citep{Schulman2017, Melo2019}.

Despite these advances, applying a single, monolithic DRL policy to simultaneously control both the humanoid's locomotion and the complex balancing manoeuvres of heavy SLs is fraught with difficulty. The vast increase in the state-action space and the potentially conflicting objectives of maintaining a stable gait while performing arm tasks can lead to intractable training processes. Hierarchical reinforcement learning approaches suggest that decomposing such complex problems can be more effective \citep{Nachum2018}, but a seamless integration remains a challenge.

\subsection{Balance Augmentation with Supernumerary Limbs}
The concept of using extra limbs for physical augmentation has been explored through various specialised strategies. One prominent approach involves using the limbs as static anchors, bracing against fixed points in the environment to provide a stable base of support. This technique is particularly effective in quasi-static scenarios like aircraft fuselage assembly, where the robot can establish a firm connection to its surroundings \citep{Parietti2014}.


For dynamic scenarios, research has focused on developing specialized appendages. These include robotic tails that modulate the body's angular momentum to counteract instabilities \citep{Abeywardena2023} and extra robotic legs that create a wider, more stable support base during locomotion with load carriage \citep{Hao2020}. While these solutions are highly effective for their specific intended functions, a generalizable strategy for maintaining dynamic walking stability in the presence of continuous, high-magnitude disturbances from multi-purpose SL arms such as the large, time-varying torques generated when rapidly repositioning a heavy payload remains an open challenge. The unpredictable, task-driven movements of these arms create a highly complex control problem that cannot be fully addressed by appendages with limited degrees of freedom or by static bracing strategies, highlighting the need for novel control frameworks \citep{Verdel2024}.

\subsection{Positioning this Research}
\label{sec:positioning_research}
This research contributes to the field by presenting a novel solution to this complex problem. Previous work on SLs for balance has often focused on static bracing or on specialized, non-anthropomorphic appendages. This project distinguishes itself by leveraging \textbf{general-purpose, anthropomorphic robotic arms for dynamic balance assistance during locomotion}. This is significant because it implies that the same limbs used for manipulation tasks could dually function as active balancing aids, greatly enhancing the versatility and utility of such a system.

The methodology also aligns with a modern trend in robotics that combines the strengths of learning-based and model-based control. While DRL provides a powerful tool for learning complex locomotion policies that are difficult to hand-engineer \citep{Peng2018}, the model-based controller offers reliability for the well-defined task of CoM regulation. This hybrid approach demonstrates a practical path forward for tackling multifaceted robotics challenges.

\section{Contribution and Objectives}

This research posits that a decoupled, hierarchical control strategy can overcome these limitations. By separating the complex problem into two more manageable layers---a low-level locomotion policy and a high-level balancing controller---it is possible to achieve stable walking for a humanoid equipped with heavy SLs. This modular approach is common in complex robotic systems, allowing for the independent development and optimization of each component, leading to a more effective overall system \citep{Sentis2010}.

The overall aim of this project is to develop and validate a novel hierarchical control framework that enables a humanoid robot to maintain stable bipedal locomotion while managing the significant \textbf{self-induced dynamic effects} imposed by heavy, articulated supernumerary limbs.

To achieve this aim, the following objectives have been established:
\begin{enumerate}
    \item To develop a walking gait for the Unitree H1 humanoid using imitation-guided Deep Reinforcement Learning, establishing a stable mobile base.
    \item To enhance the resilience of the learned locomotion policy by implementing a curriculum learning strategy that progressively introduces the mass and dynamic poses of the SLs during training.
    \item To design and implement an independent, model-based dynamic balancing controller that actively utilizes the SLs to counteract instabilities by modulating their configuration based on real-time Center of Mass (CoM) and Center of Support (CoS) feedback.
    \item To quantitatively evaluate the framework's performance in a high-fidelity physics simulation across \textbf{three interdependent and increasingly complex scenarios}, thereby validating the effectiveness of the decoupled control strategy:
    \begin{enumerate}[label=(\alph*)]
        \item \textbf{Baseline Walking:} Establish a performance benchmark with an unladen humanoid to define the characteristics of an ideal, unperturbed gait.
        \item \textbf{Static Payload:} Assess the DRL policy's ability to manage a constant, challenging load by having the humanoid walk with the SLs locked in a fixed pose.
        \item \textbf{Dynamic Balancing:} Evaluate the full hierarchical framework by activating the high-level controller to provide active balance modulation, and compare its performance against the other two scenarios.
    \end{enumerate}
\end{enumerate}

\usetikzlibrary{shapes.geometric, arrows, positioning, calc, fit, shapes.misc}
\definecolor{myblack}{HTML}{000000}
\definecolor{mydarkgray}{HTML}{4a4a4a}
\definecolor{mylightgray}{HTML}{f0f0f0}
\definecolor{myblue}{HTML}{2a6099}
\definecolor{mygreen}{HTML}{008455}
\definecolor{myorange}{HTML}{e17000}

\chapter{Methods}
\label{chap:methods}

This chapter details the hierarchical control framework, simulation environment, learning algorithms, and experimental protocols developed to achieve humanoid locomotion with supernumerary limbs (SLs). The methodology is divided into three primary components: the low-level locomotion policy trained via Deep Reinforcement Learning (DRL), the high-level model-based controller for dynamic balancing, and the experimental setup for quantitative evaluation.

\section{System Architecture}
\label{sec:system_architecture}

\subsection{Robot Model and Simulation Environment}
The study was conducted within a high-fidelity physics simulation environment to facilitate rapid prototyping and safe, extensive training. The core of the simulation is the \textbf{MuJoCo} (Multi-Joint Dynamics with Contact) physics engine, renowned for its efficiency and accuracy in simulating complex robotic systems \citep{Todorov2012}. The JAX framework was utilized for all computations to leverage its just-in-time (JIT) compilation and automatic differentiation capabilities, enabling high-performance training.

\begin{figure}[h!]
    \centering
    \includegraphics[width=0.8\textwidth]{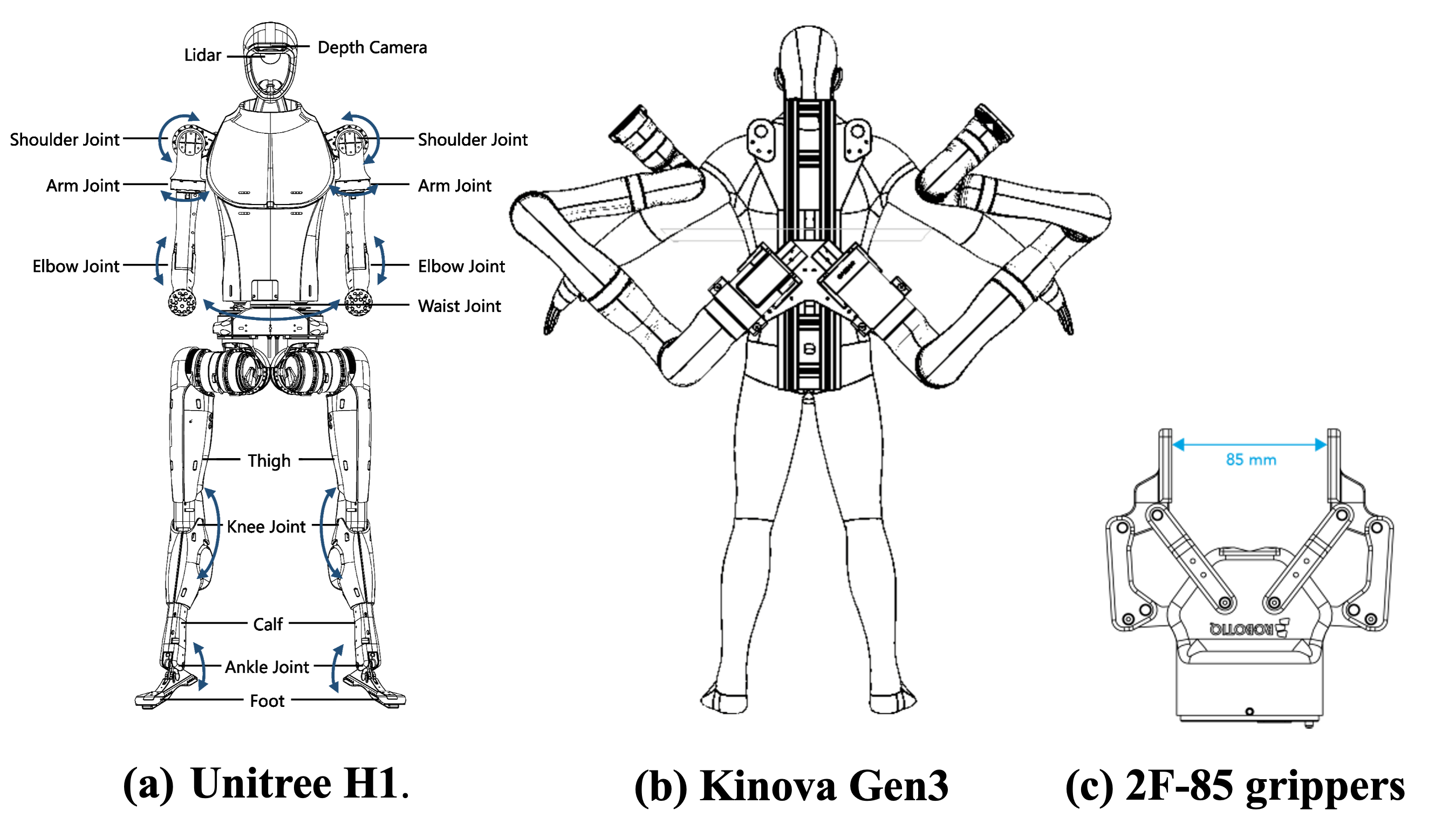}
    \caption{The composite robot model used in the simulation, illustrating (a) the Unitree H1 humanoid base, (b) the backpack-mounted Kinova Gen3 SLs, and (c) the 2F-85 grippers.}
    \label{fig:robot_model}
\end{figure}

The robotic platform, depicted in Figure~\ref{fig:robot_model}, is a composite model consisting of:
\begin{itemize}
    \item \textbf{Unitree H1 Humanoid:} A full-sized humanoid robot serving as the mobile base. Its kinematic and dynamic properties are based on the manufacturer's specifications.
    \item \textbf{Supernumerary Limbs (SLs):} Two Kinova Gen3 robotic arms are mounted on a custom backpack attached to the H1's torso. Each arm is equipped with a 2F-85 gripper.
\end{itemize}
With a total of \textbf{26 actuated degrees of freedom (DoF)}:
\begin{itemize}
    \item \textbf{H1 Humanoid (12 DoF):} The mobile base, controlling the legs and torso.
    \item \textbf{SLs (14 DoF):} Two 7-DoF Kinova Gen3 arms mounted on a backpack.
\end{itemize}
A critical detail of the simulation is that the actuator torques were \textbf{not saturated} to their physical limits. This simplification, a common practice in early-stage simulation studies, allows for focusing on the control algorithm's performance without being constrained by hardware-specific limitations. This is further discussed as a key aspect of the sim-to-real challenge in Section~\ref{sec:limitations}.

\subsection{Hierarchical Control Framework}
A decoupled, hierarchical control strategy was adopted to manage the complexity of simultaneous locomotion and balancing. This framework separates the control problem into two distinct layers, as detailed in Figure \ref{fig:control_framework}:

\vspace{-10pt}
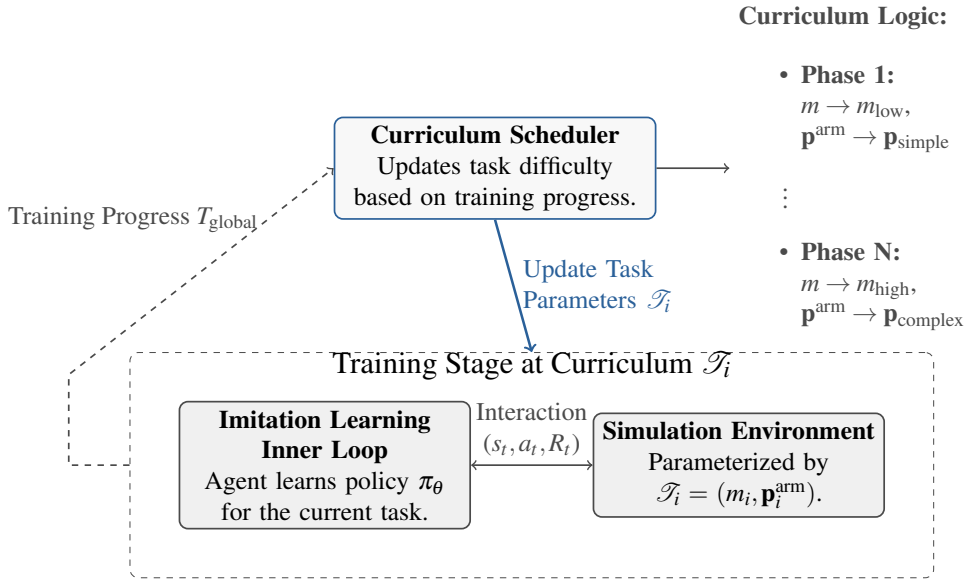
\begin{figure}[h!]
    \centering
    \scalebox{0.8}{%
    \begin{tikzpicture}[%
        node distance=1.8cm and 2cm,%
        block/.style={rectangle, draw=mydarkgray, thick, rounded corners, fill=mylightgray, minimum height=1.5cm, minimum width=3.5cm, text centered, text width=4.5cm},%
        scheduler/.style={block, draw=myblue, fill=mylightgray!50, text width=5cm, minimum width=5cm},%
        arrow/.style={->, thick, mydarkgray},%
        param_flow/.style={->, thick, myblue, line width=1.2pt},%
        progress_flow/.style={->, thick, mydarkgray, dashed}%
    ]%
    \node[scheduler] (scheduler) {\textbf{Curriculum Scheduler} \\ Updates task difficulty based on training progress.};%
    \node[block, below=3cm of scheduler, xshift=-2.8cm] (il_loop) {\textbf{Imitation Learning \\Inner Loop} \\ Agent learns policy $\pi_\theta$ for the current task.};%
    \node[block, right=of il_loop] (env) {\textbf{Simulation Environment} \\ Parameterized by $\mathcal{T}_i = (m_i, \mathbf{p}_i^{\text{arm}})$.};%
    \node[draw=mydarkgray, dashed, inner sep=0.8cm, rounded corners, fit=(il_loop) (env)] (inner_loop_box) {};%
    \node[above, font=\large, yshift=-0.6cm] at (inner_loop_box.north) {Training Stage at Curriculum $\mathcal{T}_i$};%
    \draw[param_flow] (scheduler.south) -- node[midway, right, text width=3cm] {Update Task \\Parameters $\mathcal{T}_i$} (inner_loop_box.north);%
    \draw[progress_flow] (inner_loop_box.west) -| ++(-1, 1.5) -- node[pos=0.75, left] {Training Progress $T_{\text{global}}$} (scheduler.west);%
    \draw[arrow, <->] (il_loop.east) -- node[midway, above, text width=2cm, align=center] {Interaction ($s_t, a_t, R_t$)} (env.west);%
    \node[text width=4.5cm, align=left, right=1.2cm of scheduler, mydarkgray] (curriculum_stages) {%
        \textbf{Curriculum Logic:} \\
        \begin{itemize}
            \item[\textbullet] \textbf{Phase 1:} $m \to m_{\text{low}}$, $\mathbf{p}^{\text{arm}} \to \mathbf{p}_{\text{simple}}$
            \item[\vdots]
            \item[\textbullet] \textbf{Phase N:} $m \to m_{\text{high}}$, $\mathbf{p}^{\text{arm}} \to \mathbf{p}_{\text{complex}}$
        \end{itemize}
    };%
    \draw[arrow] (scheduler.east) -- (curriculum_stages.west);%
    \end{tikzpicture}%
    }
    \caption{The overall training framework, illustrating the hierarchical structure. The outer loop consists of a \textbf{Curriculum Scheduler} that adjusts task difficulty (e.g., payload mass $m_i$ and arm pose $\mathbf{p}_i^{\text{arm}}$) based on the global training progress ($T_{\text{global}}$). The inner loop is a standard \textbf{Imitation Learning} process where the DRL agent interacts with the environment to learn a policy for the current difficulty level.}
    \label{fig:control_framework}
\end{figure}
\vspace{-10pt}

\begin{enumerate}
    \item \textbf{Low-Level Locomotion Policy:} A DRL-based policy is responsible for generating the fundamental walking gait. It controls the actuators of the humanoid's legs and lower torso, aiming to produce stable and efficient locomotion by imitating a reference motion.
    \item \textbf{High-Level Balancing Controller:} A model-based controller is dedicated to active stability augmentation. It overrides the control of the SLs, dynamically adjusting their pose to counteract perturbations and maintain the overall balance of the system.
\end{enumerate}
This decoupled approach allows the DRL agent to focus solely on mastering the core locomotion task, while the specialized balancing controller handles the complex dynamics introduced by the SLs.

\section{Low-Level Control: DRL for Locomotion}
\label{sec:drl_locomotion}

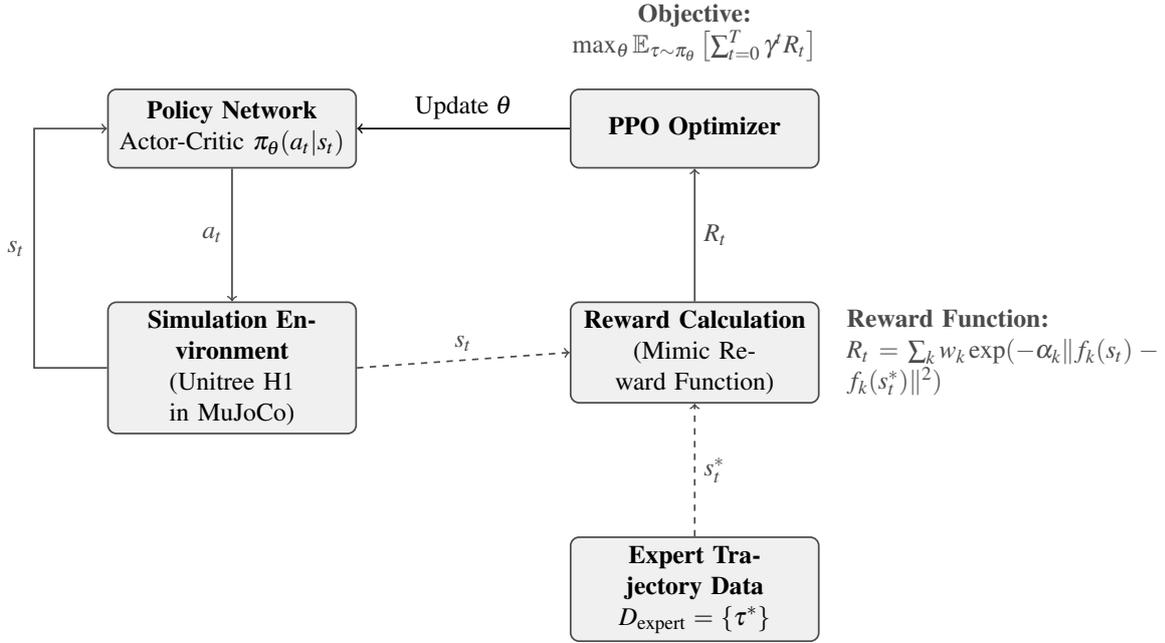
\begin{figure}[h!]
    \centering
    \scalebox{0.8}{%
    \begin{tikzpicture}[%
        node distance=2.2cm and 3.5cm,%
        block/.style={rectangle, draw=mydarkgray, thick, rounded corners, fill=mylightgray, minimum height=1.3cm, minimum width=3.8cm, text centered, text width=3.8cm},%
        arrow/.style={->, thick, mydarkgray},%
        data_flow/.style={->, thick, mydarkgray, dashed},%
        update_flow/.style={->, thick, myblack}%
    ]%
    \node[block] (policy) {\textbf{Policy Network} \\ Actor-Critic $\pi_\theta(a_t|s_t)$};%
    \node[block, below=of policy] (env) {\textbf{Simulation Environment} \\ (Unitree H1 in MuJoCo)};%
    \node[block, right=of policy] (ppo) {\textbf{PPO Optimizer}};%
    \node[block, below=of ppo] (reward) {\textbf{Reward Calculation} \\ (Mimic Reward Function)};%
    \node[block, below=of reward] (expert) {\textbf{Expert Trajectory Data} \\ $D_{\text{expert}} = \{\tau^*\}$};%
    \draw[arrow] (policy.south) -- node[midway, left] {$a_t$} (env.north);%
    \draw[arrow] (env.west) -- ++(-1.2,0) |- node[pos=0.25, left] {$s_t$} (policy.west);%
    \draw[data_flow] (env.east) -- node[midway, above] {$s_t$} (reward.west);%
    \draw[data_flow] (expert.north) -- node[midway, right] {$s_t^*$} (reward.south);%
    \draw[arrow] (reward.north) -- node[midway, right] {$R_t$} (ppo.south);%
    \draw[update_flow] (ppo.west) -- node[midway, above] {Update $\theta$} (policy.east);%
    \node[text width=5cm, align=center, above=0.3cm of ppo, mydarkgray] (ppo_formula) {\textbf{Objective:} \\ $\max_{\theta} \mathbb{E}_{\tau \sim \pi_\theta} \left[ \sum_{t=0}^{T} \gamma^t R_t \right]$};%
    \node[text width=6.5cm, align=left, right=0.3cm of reward, mydarkgray] (reward_formula) {\textbf{Reward Function:} \\ $R_t = \sum_{k} w_k \exp(-\alpha_k \| f_k(s_t) - f_k(s_t^*) \|^2)$};%
    \end{tikzpicture}%
    }
    \caption{The inner Imitation Learning loop. The \textbf{Policy Network} generates an action $a_t$ based on the current state $s_t$. The \textbf{Simulation Environment} executes this action and returns the next state. The \textbf{Reward Calculation} module compares the agent's state $s_t$ to the \textbf{Expert Trajectory} state $s_t^*$ to compute a reward $R_t$. Finally, the \textbf{PPO Optimizer} uses this reward to update the policy's parameters $\theta$.}
    \label{fig:imitation_loop}
\end{figure}

The foundation of the robot's mobility is a walking policy trained using imitation learning, based on the DeepMimic framework \citep{Peng2018}. The detailed logic of this inner training loop is shown in Figure \ref{fig:imitation_loop}.

\vspace{-10pt}
\vspace{-10pt}

\subsection{Proximal Policy Optimization (PPO)}
The policy was trained using the Proximal Policy Optimization (PPO) algorithm, a state-of-the-art DRL method known for its stability and sample efficiency \citep{Schulman2017}. PPO is an actor-critic algorithm that optimizes a clipped surrogate objective function to prevent excessively large policy updates. The objective for the policy network $\pi_{\theta}$ is to maximize the expected total discounted reward:
\begin{equation}
    \max_{\theta} \mathbb{E}_{\tau \sim \pi_{\theta}} \left[ \sum_{t=0}^{T} \gamma^t R_t \right]
    \label{eq:ppo_objective}
\end{equation}
where $\tau$ is a trajectory, $R_t$ is the reward at \mbox{timestep} $t$, and $\gamma$ is the discount factor, set to 0.99 in this work. The policy network architecture consisted of three hidden layers with [1024, 512, 256] neurons and a Tanh activation function.

\subsection{Imitation Learning and Reward Function}
To guide the learning process towards a natural, bipedal gait, an expert trajectory for \textbf{walking} was sourced from the Ubisoft La Forge Animation Dataset (LAFAN1). The agent is rewarded for mimicking this reference motion. The total reward signal $R_t$ is a weighted sum of several components, designed with the principle of "Survive First, Imitate Later":
\begin{equation}
    R_t = w_{survival}R_{survival} + w_{stability}R_{stability} + w_{imitation}R_{imitation}
\end{equation}
The imitation reward, $R_{imitation}$, further breaks down into components rewarding the similarity of joint positions, velocities, and key body site orientations relative to the expert data. The weights, detailed in Table \ref{tab:reward_weights}, were heavily skewed towards survival and stability, granting the agent the freedom to deviate from the reference motion when necessary to avoid falling, particularly under the influence of the SLs.

\begin{table}[h!]
\centering
\small
\caption{Reward function component weights.}
\label{tab:reward_weights}
\begin{tabular}{lc}
\hline
\textbf{Component} & \textbf{Weight} \\ \hline
\textit{Imitation Terms} & \\
Joint Position Match & 0.05 \\
Joint Velocity Match & 0.01 \\
Relative Site Position Match & 0.1 \\
Relative Site Quaternion Match & 0.05 \\
Relative Site Velocity Match & 0.01 \\ \hline
\textit{Stability Terms} & \\
Survival Reward (per step) & 5.0 \\
Stability (penalty for falling) & 4.0 \\ \hline
\end{tabular}
\end{table}

\subsection{Curriculum Learning}
To enable the locomotion policy to adapt to the significant disturbances from the SLs, a curriculum learning strategy was implemented. This approach gradually increases the difficulty of the task as the agent's performance improves, preventing the agent from being overwhelmed in the early stages of training. The curriculum was structured across the total 500 million training timesteps and involved two parallel difficulty ramps:
\begin{enumerate}
    \item \textbf{Payload Randomization:} The mass of the SLs' payload was gradually increased. The training started with a negligible payload, progressing in stages to the final target range of 19kg to 30kg. This allowed the agent to learn to compensate for the increasing inertial and gravitational effects.
    \item \textbf{Arm Pose Curriculum:} The static pose of the SLs was incrementally moved from a neutral, close-to-the-body position to a challenging forward-reaching posture. This curriculum was divided into four phases, each moving the arms a step closer to their final target configuration, forcing the locomotion policy to adapt to the shifting center of mass.
\end{enumerate}

\section{High-Level Control: Model-Based Dynamic Balancing}
\label{sec:dynamic_balancing}

\begin{figure}[h!]
    \centering
    \scalebox{0.8}{%
    \begin{tikzpicture}[%
        node distance=1.5cm and 2.5cm,%
        block/.style={rectangle, draw=mydarkgray, thick, rounded corners, fill=mylightgray, minimum height=1.5cm, minimum width=4.5cm, text centered, text width=4.5cm, font=\small},%
        policy_block/.style={ block, draw=mygreen, fill=mygreen!10 },%
        controller_sub_block/.style={block, draw=myorange, fill=myorange!10, minimum height=1.3cm, text width=4.2cm},%
        sum_circle/.style={circle, draw=myblack, thick, fill=white, minimum size=8mm, path picture={\draw[thick, myblack] (path picture bounding box.center) -- ++(0, 3mm) (path picture bounding box.center) -- ++(0, -3mm) (path picture bounding box.center) -- ++(3mm, 0) (path picture bounding box.center) -- ++(-3mm, 0);}},%
        arrow/.style={->, thick, mydarkgray},%
        main_flow/.style={->, thick, myblack, line width=1.2pt},%
        label/.style={font=\small\itshape}%
    ]%
    \node[block, text width=6cm] (env) {\textbf{Simulation Environment} \\ (\texttt{MjxUnitreeH1\_balance})};%
    \node[sum_circle, above=2.5cm of env] (action_mixer) {};%
    \node[policy_block, above=2.5cm of action_mixer] (policy) {\textbf{Trained Gait Policy $\pi_\theta$}};%
    \node[controller_sub_block, right=3cm of env] (state_est) {\textbf{State Estimation} \\ $\text{CoS}_{xy} = \mathbf{p}^{\text{foot}}_{xy}$ where $p^{\text{foot}}_z$ is min};%
    \node[controller_sub_block, above=0.8cm of state_est] (error_calc) {\textbf{Error Calculation} \\ $e_{xy} = \text{CoM}_{xy} - \text{CoS}_{xy}$};%
    \node[controller_sub_block, above=0.8cm of error_calc] (target_gen) {\textbf{Dynamic Target\\ Generation} \\ $q^{\text{arm}}_{\text{target}} = q^{\text{arm}}_{\text{base}} - \mathbf{K}_p^{\text{arm}} e_{xy}$};%
    \node[controller_sub_block, above=0.8cm of target_gen] (pd_control) {\textbf{PD Controller} \\ $\tau = K_p(q_{\text{target}} - q) - K_d\dot{q}$};%
    \node[draw=myorange, dashed, inner sep=0.5cm, rounded corners, fit=(state_est) (pd_control)] (balance_controller_box) {};%
    \node[above=2mm of balance_controller_box, font=\large] {Active Balance Controller};%
    \draw[main_flow] (policy.south) -- node[label, midway, left] {Humanoid Actions $a_t$} (action_mixer.north);%
    \draw[main_flow] (action_mixer.south) -- node[label, midway, left] {Combined Control $u_t$} (env.north);%
    \draw[arrow] (env.west) -- ++(-2.5, 0) |- (policy.west) node[label, pos=0.25, right] {Robot State $s_t$};%
    \draw[arrow] (env.east) -- (state_est.west) node[label, pos=0.4, above] {CoM, CoS};%
    \draw[arrow] (state_est) -- (error_calc);%
    \draw[arrow] (error_calc) -- node[label, midway, right] {$e_{xy}$} (target_gen);%
    \draw[arrow] (target_gen) -- (pd_control);%
    \draw[main_flow] (pd_control.west) -- (action_mixer.east) node[label, pos=0.3, left] {SL Arm Torques $\tau_t$};%
    \end{tikzpicture}%
    }
    \caption{Detailed control logic for the \textbf{Dynamic Balancing} scenario. The DRL \textbf{Gait Policy} generates actions for the humanoid's legs ($a_t$). In parallel, the \textbf{Active Balance Controller} uses the robot's state ($s_t$) to estimate CoM and CoS, calculates a balance error ($e_{xy}$), and generates compensatory torques for the SL arms ($\tau_t$) via a PD controller. These two control signals are combined and sent to the simulation environment.}
    \label{fig:balancing_logic}
\end{figure}
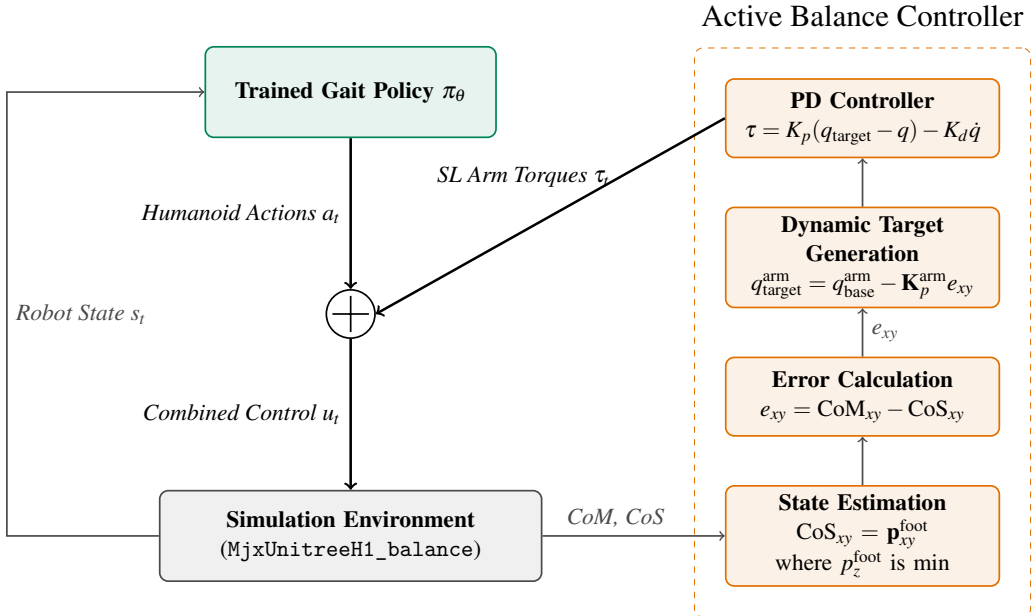

While the DRL policy learns to walk, the high-level controller actively uses the SLs to provide dynamic stability. This controller operates independently of the DRL agent and is based on a simplified rigid-body dynamics model. The logic for this controller is detailed in Figure \ref{fig:balancing_logic}.

\subsection{State Estimation for Balancing}
\label{sec:state_estimation}
The high-level controller's function relies on real-time estimation of the robot's balance state. This is achieved using two key metrics calculated at each timestep:
\begin{enumerate}
    \item \textbf{Center of Mass (CoM):} The total body CoM is calculated as the weighted average of the positions of all individual body links.
    \item \textbf{Center of Support (CoS):} To achieve a physically accurate stability reference, the Center of Support is calculated based on the ground contact forces and the support polygon. The calculation adapts to the phase of the gait:
    \begin{itemize}
        \item During the \textbf{single-support phase}, the CoS is defined as the geometric center of the stance foot's contact area.
        \item During the \textbf{double-support phase}, the controller first calculates the Center of Pressure (CoP) for each foot using the measured ground reaction forces ($F_z$) and moments ($T_x, T_y$). The global CoS is then computed as the force-weighted average of the individual CoPs.
    \end{itemize}
    This is formally expressed as:
    \begin{equation}
        \mathbf{CoS}_{xy} =
        \begin{cases}
            \mathbf{p}_{\text{stance\_foot}, xy} & \text{if single support} \\
            \frac{\mathbf{CoP}_{\text{L}} F_{z,\text{L}} + \mathbf{CoP}_{\text{R}} F_{z,\text{R}}}{F_{z,\text{L}} + F_{z,\text{R}}} & \text{if double support}
        \end{cases}
    \end{equation}
    where the per-foot CoP is given by $\mathbf{CoP} = [-T_y/F_z, T_x/F_z]$ in the foot's local frame. This physically-grounded approach ensures that the stability target is accurately represented throughout the entire gait cycle, including the critical transitions between single and double support phases.
\end{enumerate}
During dynamic walking, a controlled misalignment between the CoM and CoS is essential for forward propulsion. Therefore, the vector difference, $\mathbf{d}_{xy} = \mathbf{CoM}_{xy} - \mathbf{CoS}_{xy}$, is not treated as a static error to be nullified. Instead, it serves as a \textbf{dynamic stability indicator}.

\subsection{Balancing Controller and Control Fusion}
The balancing controller implements a reactive strategy, using the dynamic stability indicator $\mathbf{d}_{xy}$ to adjust the target joint angles of the two SLs. The objective is to move the arms in a way that generates compensatory momentum, shifting the total body CoM back towards a stable region relative to the CoS. The target arm pose, $\mathbf{q}_{\text{target}}^{\text{arm}}$, is calculated by modulating a constant, neutral base pose, $\mathbf{q}_{\text{base}}^{\text{arm}}$, with the stability indicator. This base pose defines a "home" configuration where the arms are held slightly forward and down, and it remains fixed throughout the dynamic balancing trials. The modulation is scaled by a gain matrix $\mathbf{K}_{\text{p}}^{\text{arm}}$:
\begin{equation}
    \mathbf{q}_{\text{target}}^{\text{arm}} = \mathbf{q}_{\text{base}}^{\text{arm}} - \mathbf{K}_{\text{p}}^{\text{arm}} \mathbf{d}_{xy}
\end{equation}
This linear, heuristic control law was chosen for its computational simplicity and real-time feasibility. While an optimal target pose could be computed by solving a non-linear, whole-body optimization problem, such an approach would be computationally expensive. The goal here is not to find a single, perfect corrective pose, but to provide continuous, high-frequency dynamic damping, for which this proportional strategy proves effective. It is important to note that the fixed pose used for the "Static Payload" experimental scenario is a more challenging, forward-reaching posture, distinct from the neutral $\mathbf{q}_{\text{base}}^{\text{arm}}$ used here.

A Proportional-Derivative (PD) controller then calculates the required torques $\boldsymbol{\tau}_{\text{arm}}$ to drive the SL arm joints towards this dynamic target:
\begin{equation}
    \boldsymbol{\tau}_{\text{arm}} = \mathbf{K}_p (\mathbf{q}_{\text{target}}^{\text{arm}} - \mathbf{q}_{\text{current}}^{\text{arm}}) - \mathbf{K}_d \dot{\mathbf{q}}_{\text{current}}^{\text{arm}}
    \label{eq:pd_control}
\end{equation}
The gain matrices for both the target modulation ($\mathbf{K}_{\text{p}}^{\text{arm}}$) and the PD tracking controller ($\mathbf{K}_p$, $\mathbf{K}_d$) were determined empirically through an iterative tuning process within the simulation. Starting with low values, the gains were manually adjusted by observing the system's response. The objective was to achieve a critically damped behavior, where the arms responded quickly and decisively to balance perturbations without introducing significant overshoot or oscillation that could further destabilize the humanoid. This manual tuning is a standard practice for tuning low-level controllers where a precise analytical model for gain selection is unavailable or impractical.

These calculated arm torques are then fused with the output of the DRL policy. In each simulation step, the torques for the SL arm actuators are overridden by the output of the balancing controller ($\boldsymbol{\tau}_{\text{arm}}$), while the torques for the humanoid's leg and torso actuators are taken directly from the DRL policy's action. This clean separation ensures that each controller operates within its designated domain.

\section{Experimental Evaluation}
\label{sec:experimental_evaluation}
To validate the effectiveness of the hierarchical framework, a quantitative analysis was performed across three distinct experimental scenarios.
\begin{enumerate}
    \item \textbf{Baseline Walking:} The humanoid walks without the SL backpack, controlled solely by the DRL policy trained on a base curriculum (no payload, neutral arms). This establishes the benchmark for an unperturbed gait.
    \item \textbf{Static Payload:} The humanoid walks with the SLs locked in a fixed, forward-reaching pose. The DRL policy used here was fully trained with the complete curriculum, but the high-level balancing controller is disabled. This isolates the effect of the learned policy against a constant, challenging load.
    \item \textbf{Dynamic Balancing:} The full hierarchical framework is active. The fully trained DRL policy controls locomotion while the high-level controller dynamically actuates the SLs for balance.
\end{enumerate}
Key performance metrics were defined to assess dynamic stability, including CoM trajectory similarity, stability recovery based on CoM-CoS distance, and bipedal coordination via Ground Reaction Force (GRF) analysis. The phase-plane analysis of Ground Reaction Forces, detailed in Section~\ref{sec:grf_analysis}, involved fitting an ellipse to the data points for each scenario. This fitting was performed using Principal Component Analysis (PCA). The first principal component determines the direction of the major axis of the ellipse, representing the primary axis of variance in the coordinated forces. The "Orientation Error" was then calculated as the angle between this major axis and the ideal 135-degree anti-phase axis. These metrics are detailed further in the Results chapter.
\chapter{Results}
\label{chap:results}

This chapter presents the results of the training process and the quantitative evaluation of the hierarchical control framework. The findings are organized into four sections: DRL training performance, Center of Mass (CoM) trajectory analysis, dynamic balance modulation, and an exploratory analysis of gait coordination.

\section{DRL Training Performance}
\label{sec:training_performance}
\begin{figure}[h!]
    \centering
    \includegraphics[width=\textwidth]{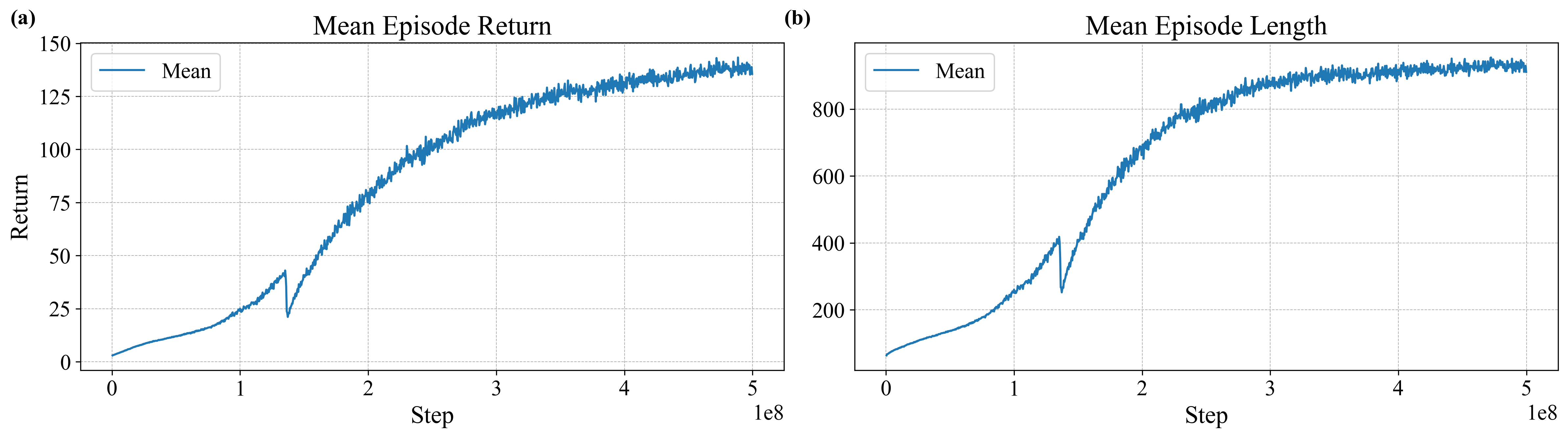}
    \caption{Training performance of the PPO agent over 500 million environment steps. (a) Mean Episode Return. (b) Mean Episode Length. The agent's consistent improvement, punctuated by temporary dips aligned with curriculum changes, demonstrates successful adaptation.}
    \label{fig:training_curves}
\end{figure}

The DRL agent was trained for a total of 500 million environment steps. The curriculum learning strategy progressively increased the task difficulty by increasing the payload mass and adjusting the SL arm poses at intervals of 100 million steps. The agent's learning progress was monitored via the mean episode return (task achievement) and mean episode length (stability). Figure~\ref{fig:training_curves} shows that both metrics trended consistently upwards, demonstrating the agent's increasing performance. The periodic dips, particularly around the 100-million-step marks, correspond to the scheduled increases in curriculum difficulty. The agent's ability to quickly recover and continue improving after each increase validates the effectiveness of the curriculum strategy in adapting the policy to the challenging dynamics of the SLs.

\section{Center of Mass Trajectory Analysis}
\label{sec:com_analysis}
To assess how the different control strategies affected the overall walking pattern, the trajectory of the robot's Center of Mass (CoM) was recorded for each of the three experimental scenarios. The "Baseline Walking" scenario serves as the reference for an ideal, unperturbed gait. To compare the "Static Payload" and "Dynamic Balancing" scenarios against this baseline, their similarity was quantified using the Dynamic Time Warping (DTW) distance. DTW is a metric for measuring similarity between two temporal sequences that may vary in speed, providing a single value where a lower number indicates a higher degree of similarity in the dynamic pattern.

\begin{figure}[h!]
    \centering
    \includegraphics[width=\textwidth]{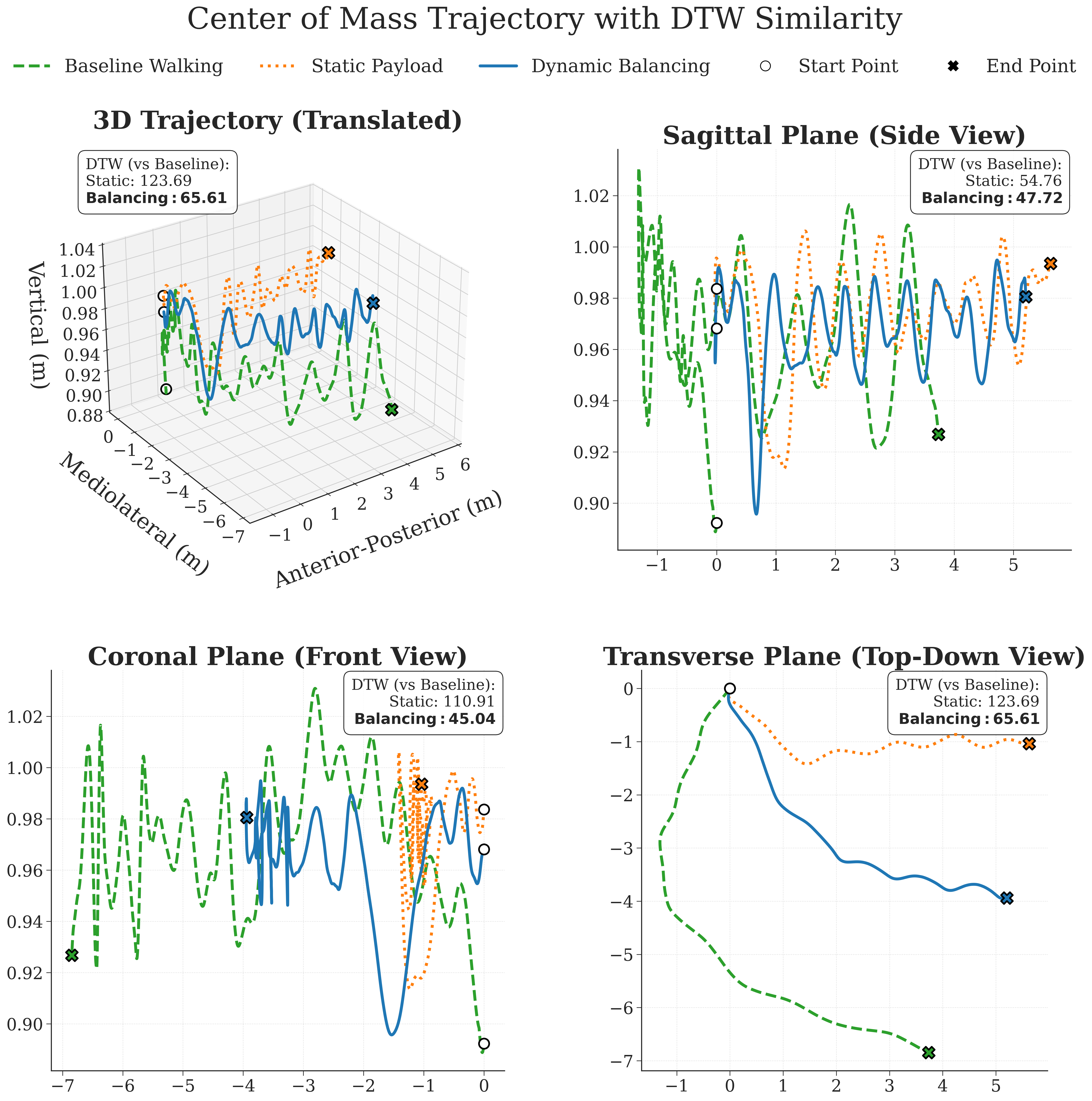}
    \caption{CoM trajectories for the three scenarios. DTW distances are relative to the "Baseline Walking" trajectory. The lower DTW score for "Dynamic Balancing" indicates its dynamic pattern and rhythm are better preserved.}
    \label{fig:com_trajectory_final}
\end{figure}

The analysis, presented in Figure~\ref{fig:com_trajectory_final}, reveals that the "Dynamic Balancing" strategy results in a CoM trajectory with a DTW distance of 65.54 from the baseline, a significant reduction of approximately 47\% compared to the "Static Payload" trajectory's distance of 123.71. This lower DTW score indicates that the active balancing controller is effective at mitigating the high-frequency disturbances introduced by the SLs. Although the absolute paths of all trajectories diverge over the course of the trial, particularly in the transverse plane, the dynamic balancing controller better preserves the fundamental dynamic characteristics and rhythm of the original unladen gait. In contrast, the static payload introduces more severe, uncompensated disturbances, causing a greater deviation in both the CoM's dynamic pattern and its final position.

\section{Analysis of Dynamic Balance Modulation}
\label{sec:balance_modulation}
Walking is inherently a process of controlled instability, requiring continuous modulation of the body's balance. A key indicator of this dynamic state is the distance between the horizontal projection of the Center of Mass and the Center of Support (CoM-CoS distance). This distance naturally oscillates throughout the gait cycle: it increases during the swing phase as the body "falls" forward and decreases after foot-strike as the body re-stabilizes over the new support foot. The controller's objective is not to eliminate these crucial oscillations, but to effectively manage their magnitude in the presence of disturbances.

\begin{figure}[h!]
    \centering
    \includegraphics[width=\textwidth]{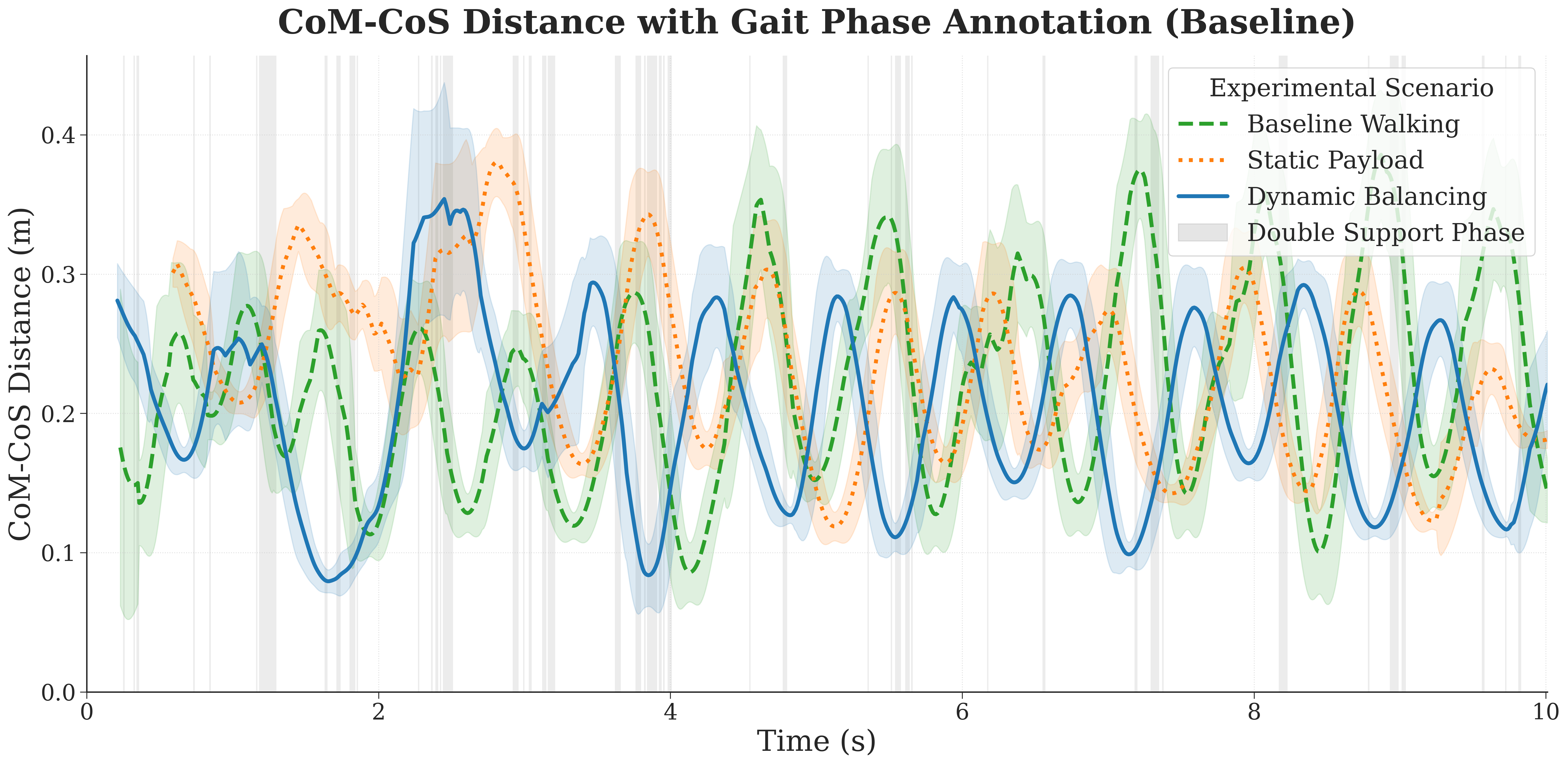}
    \caption{A qualitative overview of the CoM-CoS distance oscillations over a 10-second trial. For the baseline, shaded regions indicate double-support phases, illustrating the cyclical nature of balance modulation.}
    \label{fig:com_cos_distance_time_final}
\end{figure}

Figure~\ref{fig:com_cos_distance_time_final} provides a qualitative overview of these oscillations. A key observation is that the frequency of the double-support phase (shaded regions) is not perfectly constant, even in the baseline scenario. This is an emergent behavior of the DRL policy. Unlike a traditional controller with a fixed cadence, the DRL agent's primary objective is stability. It continuously makes subtle timing adjustments to the gait cycle to maintain balance, which results in a stable but not perfectly rhythmic walking pattern. This also explains the slight drift in the baseline's oscillation amplitude, which is an artifact of this adaptive control strategy, not an intended circular path.

\begin{figure}[h!]
    \centering
    \includegraphics[width=\textwidth]{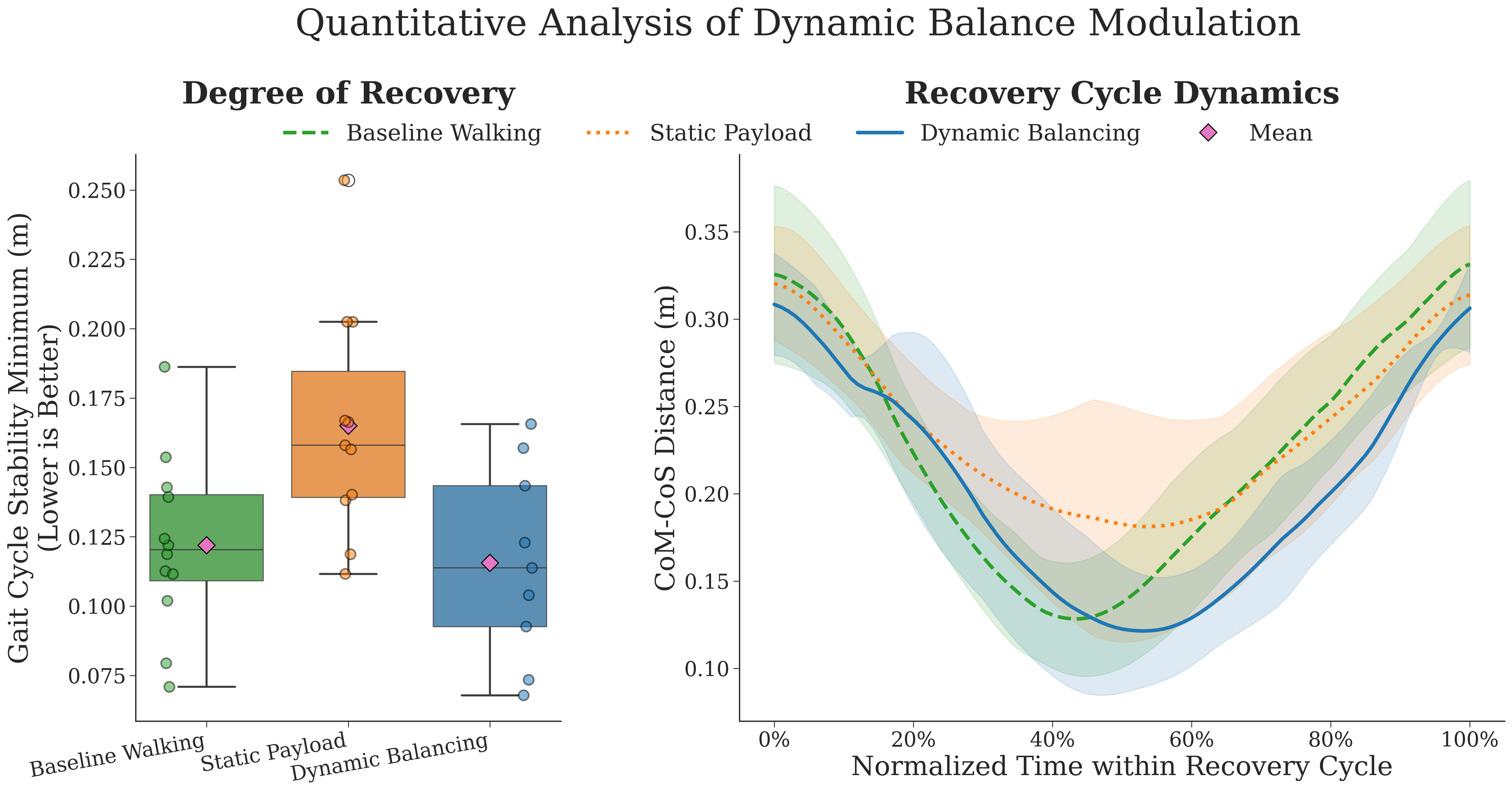}
    \caption{Quantitative analysis of dynamic balance modulation. \textbf{Left:} Boxplot of the GCSM. Lower values indicate more complete re-stabilization. Diamond markers indicate the mean. \textbf{Right:} Average CoM-CoS distance over a normalized recovery cycle (from peak instability to peak recovery). The method for normalizing and averaging individual cycles is described in the text.}
    \label{fig:stability_analysis_final}
\end{figure}

For a rigorous comparison, each gait cycle (from one peak of CoM-CoS distance to the next) was isolated and time-normalized, as shown in Figure~\ref{fig:stability_analysis_final} (Right). This ensures that corresponding phases of the gait cycle are directly compared across all scenarios. To quantify the effectiveness of re-stabilization within each cycle, we define a new metric: the \textbf{Gait Cycle Stability Minimum (GCSM)}. This metric represents the minimum CoM-CoS distance achieved during the recovery phase of the i-th gait cycle:
\begin{equation}
    \text{GCSM}_i = \min_{t_{\text{peak},i} < t < t_{\text{peak},i+1}} D(t)
    \label{eq:gcsm}
\end{equation}
where $D(t)$ is the CoM-CoS distance at time $t$, and $t_{\text{peak},i}$ is the time of the $i$-th peak. The GCSM quantifies how successfully the system arrests the forward fall and achieves a stable state over the stance foot. A lower GCSM value indicates a more effective and complete recovery. This metric was chosen over an average distance because it specifically targets the critical moment of maximum recovery, providing a more direct measure of the controller's ability to handle the cycle's peak instability.

The aggregated results, shown in Figure~\ref{fig:stability_analysis_final} (Left), highlight a clear performance benefit for the "Dynamic Balancing" scenario. The boxplot of GCSM values (left) shows that "Dynamic Balancing" achieves the lowest median value, signifying consistently more effective re-stabilization after each step. Furthermore, the plot of the averaged, normalized recovery cycle dynamics (right) confirms this finding, showing that the "Dynamic Balancing" curve reaches a significantly lower trough than the other scenarios. This demonstrates that the active SL controller enables the robot to better manage the dynamic fluctuations inherent in walking, leading to a more stable state at the most critical point of recovery within each gait cycle.

\section{Exploratory Analysis of Gait Coordination}
\label{sec:grf_analysis}
To explore how the different loading conditions influenced the coordination of the bipedal gait, a phase-plane analysis of the vertical Ground Reaction Forces (GRF) was conducted. In an ideal gait, the GRFs of the left and right feet exhibit a perfect anti-phase relationship, resulting in a data distribution oriented at 135°. The deviation from this ideal, termed "Orientation Error," can serve as an indicator of gait coordination.

\begin{figure}[h!]
    \centering
    \includegraphics[width=\textwidth]{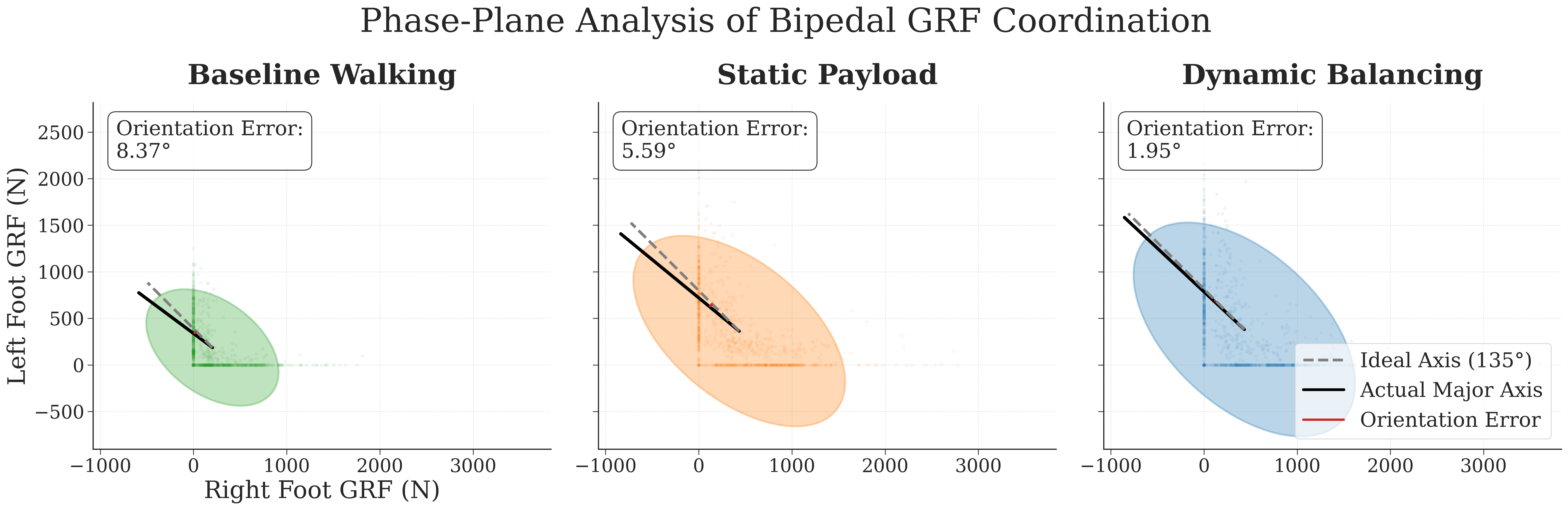}
    \caption{Phase-plane analysis of vertical GRF. "Orientation Error" measures the deviation of the ellipse's major axis from the ideal 135° anti-phase axis. This plot illustrates relative differences in bipedal coordination.}
    \label{fig:grf_analysis_final}
\end{figure}

It is important to note that the baseline gait, learned through imitation, is not perfectly optimal and exhibits a benchmark Orientation Error of 8.37°. This inherent sub-optimality may stem from imperfections in the reference motion capture data or from the DRL policy prioritizing stability over perfect mimicry. Therefore, this analysis focuses not on achieving a perfect score, but on observing the \textit{relative changes} in coordination across the different experimental conditions, as shown in Figure~\ref{fig:grf_analysis_final}.

Interestingly, the addition of a "Static Payload" was observed to reduce the error to 5.59°. A plausible physical explanation is that the mass of the SLs, held in a fixed sloping-downward pose, lowered the robot's overall center of mass. A lower CoM inherently increases the system's passive stability, which may have allowed the DRL policy to execute a more coordinated gait pattern under this constant, predictable load. The "Dynamic Balancing" scenario achieved the lowest error among the three conditions at 1.95°. This trend suggests that by offloading the primary balancing task to the high-level controller, the low-level locomotion policy can better adhere to its learned coordination pattern. However, while this result is consistent with the study's main findings, a more rigorous statistical analysis would be required to confirm the significance of these coordination differences.

\chapter{Discussion}
\label{chap:discussion}

This study successfully developed and validated a hierarchical control framework for a humanoid robot equipped with supernumerary limbs (SLs). The results presented in the previous chapter demonstrate that by decoupling the control of locomotion and dynamic balancing, the system can achieve stable walking even when subjected to the significant dynamic perturbations of the SLs. This chapter interprets these findings, discusses their implications in the context of existing literature, acknowledges the limitations of the current work, and proposes directions for future research.

\section{Interpretation of Key Findings}
\label{sec:interpretation}
The quantitative results from the three experimental scenarios provide valuable insights into the performance of the proposed hierarchical approach, highlighting both its benefits and the complexities of the control problem.

First, the successful training of the low-level locomotion policy confirms the viability of using imitation learning with a carefully structured curriculum. The learning curves (Figure~\ref{fig:training_curves}) show that the DRL agent developed a policy capable of handling the significant, constant load imposed by the SLs during evaluation. This capability was achieved through a curriculum that adapted the policy by progressively introducing challenges during training, namely increasing payload mass and more destabilizing arm poses. This adaptability during training is the foundation upon which the entire hierarchical system is built.

Second, the analysis of the Center of Mass trajectory (Figure~\ref{fig:com_trajectory_final}) clarifies the role of the dynamic balancing controller. The ideal objective for such a system is to maintain stability for any given trajectory. While the controller does not force the robot's trajectory to perfectly match the unladen baseline, the significantly lower DTW distance under active balancing is a key finding. It suggests that the high-level controller helps preserve the fundamental dynamic characteristics and rhythm of the learned gait better than in the static payload case. The data indicates that the active controller partially mitigates the corrupting influence of the SLs on the gait pattern being executed by the low-level DRL policy.

Third, the analysis of dynamic balance modulation (Figure~\ref{fig:stability_analysis_final}) provides the most direct evidence of the controller's function in enhancing intra-cycle stability. The lower Gait Cycle Stability Minimum (GCSM) achieved in the dynamic balancing scenario demonstrates how the SLs can be transformed from a purely destabilizing liability into a functional asset for active re-stabilization within each step. They are actively used to achieve a more stable state at the most critical point of the gait cycle, showcasing a clear functional benefit for augmenting a humanoid with actively controlled limbs.

Finally, the exploratory phase-plane analysis of Ground Reaction Forces (Figure~\ref{fig:grf_analysis_final}) offers a tentative insight into a possible synergistic relationship between the control layers. The trend towards a lower Orientation Error suggests that by offloading the primary balancing task to the high-level SL controller, the low-level locomotion policy is able to better adhere to its learned coordination pattern. This finding, while requiring further statistical validation, supports the core hypothesis of the decoupled framework: that separating control responsibilities can lead to measurable performance benefits in specific aspects of the task.

\section{Limitations of the Study}
\label{sec:limitations}
This study has several limitations that must be acknowledged.

First, the most significant limitation is the reliance on a simulation environment. While MuJoCo provides high-fidelity physics, the infamous "sim-to-real" gap remains a substantial hurdle. A critical aspect of this gap is that the simulation did not enforce the torque limits of the Unitree H1's motors. It is therefore uncertain whether the physical robot could support the static weight of the SL payload or generate the required compensatory torques without violating hardware constraints. Transferring this system to the physical hardware would require not only substantial effort in domain randomization but also a thorough validation of the required actuator torques against the robot's physical capabilities.

Furthermore, the current framework treats balancing as the sole function of the SLs. The reason for not integrating manipulation tasks was to manage the project's scope and focus on solving the foundational problem of maintaining stability. Integrating task-space objectives for the arms would introduce a complex multi-objective optimization problem, requiring the controller to continuously arbitrate between the often-competing demands of balance maintenance and task execution. While this integration is the ultimate goal for such a system, it was deemed beyond the scope of this initial investigation, which aimed to first establish a viable baseline for stable locomotion under heavy, dynamic loads. 

Finally, the scope of the analysis itself represents a limitation. While the results effectively demonstrate that the balancing controller improves stability (e.g., by reducing GCSM), the study did not deeply investigate how it achieves this at the joint level. A detailed, moment-to-moment analysis of the SLs' emergent motion strategies was not performed, limiting the current depth of interpretation regarding the controller's specific corrective actions.

\section{Future Work}
\label{sec:future_work}
The findings and limitations of this project suggest several promising avenues for future research.

The most immediate and critical next step is to address the sim-to-real transfer challenge. This would involve deploying the hierarchical controller on the physical robotic platform, which requires a thorough validation of actuator torque requirements against the hardware's physical limits.

Prior to developing a more complex controller, a deeper analysis of the current system's emergent behavior is warranted. Future work should visualize the SLs' joint angle trajectories, time-locked to the humanoid's gait cycle. Correlating how the arms move angularly with the CoM-CoS distance oscillations would provide crucial insights into how the reactive controller impacts the correctness of the gait on a moment-to-moment basis. This analysis would help identify the specific strategies the arms employ to provide stabilization and reveal the limitations of the current heuristic approach.

Finally, the insights from this detailed motion analysis would then directly inform the development of a more sophisticated, unified controller. Such a controller could use optimization-based techniques to simultaneously solve for manipulation task goals and balance constraints, treating the SLs' contribution to stability as a component within a whole-body control objective. This would move the system from a decoupled hierarchy to a fully integrated control architecture, realizing the full potential of a humanoid robot augmented with functional supernumerary limbs.

\chapter{Conclusion}
\label{chap:conclusion}

This thesis presented a novel hierarchical control framework to address the significant challenge of maintaining stable bipedal locomotion for a humanoid robot augmented with heavy, actuated supernumerary limbs (SLs). The core of this work was the strategic decoupling of control responsibilities, in which a low-level policy trained with Deep Reinforcement Learning (DRL) managed the fundamental walking gait, while a high-level model-based controller dynamically utilized SLs for active balance. 

Through a comprehensive set of experiments conducted in a physics-based simulation, this study provided valuable insights into the performance of this decoupled approach. The key contributions and findings are summarized as follows:
\begin{enumerate}
    \item A locomotion policy was successfully trained using imitation learning in conjunction with a curriculum strategy. This policy proved capable of adapting to the substantial and progressively increasing dynamic perturbations imposed by the SLs.
    \item The active balancing controller demonstrated its ability to preserve the underlying gait pattern more effectively than a static payload condition. This was evidenced by a significantly lower Dynamic Time Warping (DTW) distance between the robot's CoM trajectory and the unperturbed baseline, indicating a better preservation of the gait's dynamic characteristics.
    \item The framework enhanced the robot's ability to modulate its dynamic balance within each gait cycle. By actively using the SLs for re-stabilization, the system consistently achieved a more stable state at the most critical point of recovery, transforming the limbs from a simple payload into a functional asset.
    \item A potential synergistic relationship between the control layers was observed. Exploratory analysis suggested that by offloading the primary balancing task, the locomotion policy was able to execute a more coordinated bipedal pattern, though this finding requires further statistical validation.
\end{enumerate}

In conclusion, this work validates that a decoupled, hierarchical strategy is a viable method for managing the immense complexity of a humanoid-SL system. The quantitative results show measurable improvements in gait pattern preservation and intra-cycle stability.

\renewcommand{\bibname}{Reference}
\bibliographystyle{dcu}  
\bibliography{reference}

\end{document}